\documentclass[a4paper,12pt]{article}
\usepackage[utf8]{inputenc}
\usepackage{lineno,hyperref}

\usepackage{textcomp}
\usepackage{array}
\usepackage{amsmath,amssymb} 
\usepackage{color}
\usepackage{diagbox, multirow}
\usepackage{makecell}
\usepackage{graphicx}
\usepackage{latexsym}
\usepackage{xcolor}
\usepackage{authblk}
\usepackage{amsmath,amssymb}
\makeatletter
\newsavebox\myboxA
\newsavebox\myboxB
\newlength\mylenA

\newcommand*\xoverline[2][0.75]{%
	\sbox{\myboxA}{$\m@th#2$}%
	\setbox\myboxB\null
	\ht\myboxB=\ht\myboxA%
	\dp\myboxB=\dp\myboxA%
	\wd\myboxB=#1\wd\myboxA
	\sbox\myboxB{$\m@th\overline{\copy\myboxB}$}
	\setlength\mylenA{\the\wd\myboxA}
	\addtolength\mylenA{-\the\wd\myboxB}%
	\ifdim\wd\myboxB<\wd\myboxA%
	\rlap{\hskip 0.5\mylenA\usebox\myboxB}{\usebox\myboxA}%
	\else
	\hskip -0.5\mylenA\rlap{\usebox\myboxA}{\hskip 0.5\mylenA\usebox\myboxB}%
	\fi}
\makeatother
\usepackage{mathtools, geometry}

\usepackage[caption=false,font=footnotesize]{subfig}
\usepackage{algorithmic}
\usepackage[ruled,vlined]{algorithm2e}

\begin{document}

\title{Network Fission Ensembles for Low-Cost Self-Ensembles}


\author[1]{Hojung Lee}
\author[1]{Jong-Seok Lee\thanks{Corresponding author: \texttt{jong-seok.lee@yonsei.ac.kr}}} 
\affil[1]{School of Integrated Technology, Yonsei University, Incheon 21983, South Korea}

\maketitle

\begin{abstract}
Recent ensemble learning methods for image classification have demonstrated improved accuracy with low extra cost. 
However, they still rely on multiple trained models for ensemble inference, which can become a significant burden as the model size grows.
Moreover, their performance has been somewhat limited compared to Deep Ensembles, primarily due to the lower performance of individual ensemble members.
In this paper, we propose a low-cost ensemble learning and inference method called Network Fission Ensembles (NFE), which transforms a conventional network into a multi-exit structure allowing predictions to be made at different stages and enabling ensemble learning. 
To achieve this, we group the weight parameters in the layers into several sets and create multiple auxiliary paths by combining each set to construct multi-exits. 
We call this process Network Fission. 
Since this process simply changes the existing network structure to have multiple exits (i.e., classification outputs) without using additional networks, there is no extra computational burden. 
Furthermore, we employ an ensemble knowledge distillation technique exploiting the losses of all exits to train the network, so that we can achieve high generalization performance despite the reduced network size of each path composed of pruned weights. 
With our simple yet effective method, we achieve an accuracy of 83.5\% on CIFAR100 with Wide-ResNet28-10, surpassing the best existing ensemble method, Deep Ensembles, which achieves 83.0\%, while only one-third of the computational complexity is required in our method. The code is available at https://github.com/hjdw2/NFE.

\end{abstract}

\section{Introduction}
\label{sec:intro}
Forming an ensemble of multiple neural networks is one of the easiest and most effective ways to enhance the performance of deep neural networks for image classification. 
Simple combination, such as majority vote or averaging the outputs of multiple models, Deep Ensembles \cite{ens3}, can significantly improve the generalization performance \cite{ens5}. 
Despite its simplicity, however, it is difficult to use such a method when the model size or the number of training data increases since it requires a significant amount of computational resources. 

In order to overcome this limitation, several ensemble learning methods with low extra computational costs have recently been proposed.
To reduce the training burden of ensemble learning, these methods obtain multiple trained models using efficient approaches, such as exploiting smaller pruned sub-networks instead of intact original networks \cite{PAT,Freeticket}, obtaining multiple models through a single learning process using a cyclic learning rate schedule \cite{Snapshot,FGE}, training a single network structure that embeds multiple ensemble members \cite{Batchens,MIMO}, or utilizing grouped convolution in a conventional network to provide multiple outputs \cite{GEnet, PACKED}.
Additionally, some methods perform the operation of Deep Ensembles in a cascade manner to make it more efficient \cite{WBE}.
While previous methods have successfully reduced computational costs, most of them still rely on multiple models for ensemble inference. 
Therefore, as the model size increases, the additional computational cost multiplies with the number of models. Furthermore, these methods typically underperform compared to Deep Ensembles in terms of accuracy.

\begin{figure}[t] 
	\centerline{\includegraphics[width=0.7\textwidth]{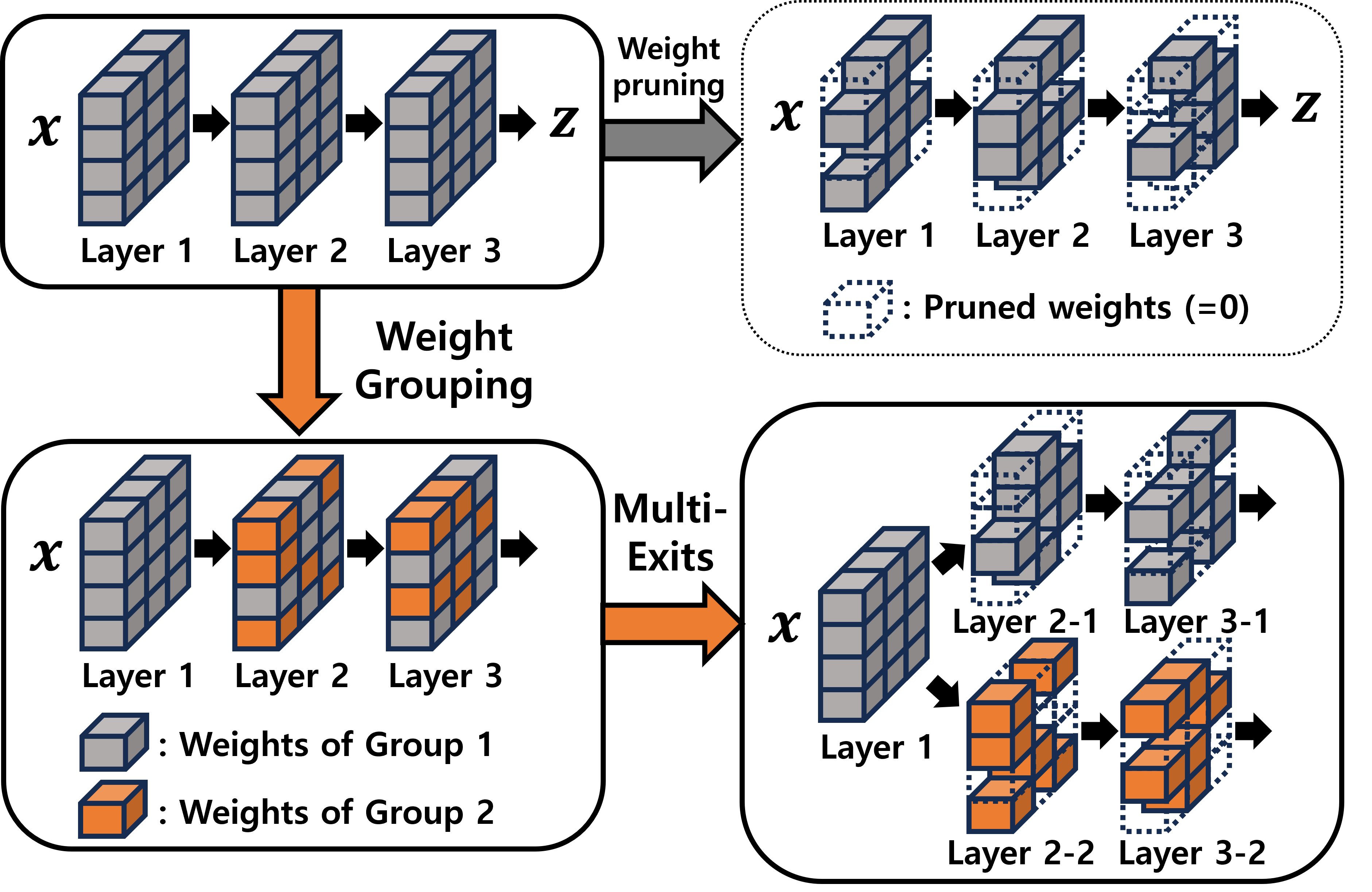}}
	\caption{Illustration comparing common weight pruning with the proposed Network Fission process. The small boxes represent individual weight parameters. We group the weights of layers into several sets and create multiple auxiliary classifier paths (exits) to construct multi-exits. This Network Fission enables us to obtain multiple outputs with a single network for ensemble learning at almost zero cost.}
	\label{Figure Intro}
\end{figure}

To address these issues, we focus on two key aspects: first, using only a single model, and second, developing a training method that enhances ensemble performance. We propose a new ensemble learning method that can produce multiple outputs with a single conventional network (e.g., ResNet \cite{resnet}) without additional modules and enhance the ensemble performance by training the model using ensemble knowledge distillation. Most current CNN architectures are over-parameterized, i.e., have more weight parameters than necessary to learn the given training data. Thus, even if some of the weights are pruned, it has little to no impact on performance. Based on this fact, we propose to prune weights and attach the pruned weights as auxiliary paths to the network, creating a multi-exit structure. 
The multi-exit structure, as demonstrated in previous studies \cite{branchnet,kd1,MATE}, is a method that improves the overall performance of the network through regularization by optimizing the joint loss across multiple exits, ultimately achieving high performance.
To implement this, we group the weight parameters in each stage and create multiple auxiliary paths combining each of them as shown in Figure \ref{Figure Intro}, which we call Network Fission. 
This transformation does not use additional networks for both training and inference but only changes the network structure (computation process) using the existing weights. Therefore, this approach avoids increasing the number of parameters or memory usage, enabling ensemble learning and inference at nearly zero additional cost and effectively addressing the first issue. 
Next, to enhance the ensemble performance, we use ensemble knowledge distillation.
When the multi-exits are properly designed so that the performance of each exit is comparable, ensembling the outputs from all exits can help compensate for decisions that individual exits make with low confidence, leading to improved ensemble performance compared to individual outputs \cite{EKD, BEED}.
Therefore, using this ensemble output as the knowledge distillation teacher and applying the distillation loss to all exits can significantly improve the performance of the network, even compensating for the weaker performance of pruned paths.
As a result, this approach can achieve high ensemble performance and effectively addresses the second issue.
After training the multi-exits, the outputs of all exits are combined for ensemble inference.
Thus, ensemble inference can be performed only with one network without the need to load multiple models.
We call our method Network Fission Ensembles (NFE).
Moreover, to further reduce the training burden, we can apply the pruning-at-initialization (PaI) method before training.
We show that NFE achieves highly satisfactory performance at a low cost and is effective compared to other state-of-the-art ensemble methods.

To sum up, our contributions are as follows: 
\begin{itemize}
	\item Single model ensembles: Our method uses a single model for ensemble learning and inference, so even if the model size increases, there is no additional computational overhead associated with ensemble learning.
	\item Ensemble knowledge distillation: Through ensemble knowledge distillation, our network, even with a multi-exit design consisting of pruned weights, can be trained to achieve high overall performance.
	\item High performance with low cost: We demonstrate the superiority of our method by comparing it with recent state-of-the-art low-cost ensemble learning methods as well as Deep Ensembles, showing that it achieves improved classification performance but incurs only the computational burden of a single model.
\end{itemize}

\begin{figure*}[t] 
	\centerline{\includegraphics[width=0.95\textwidth]{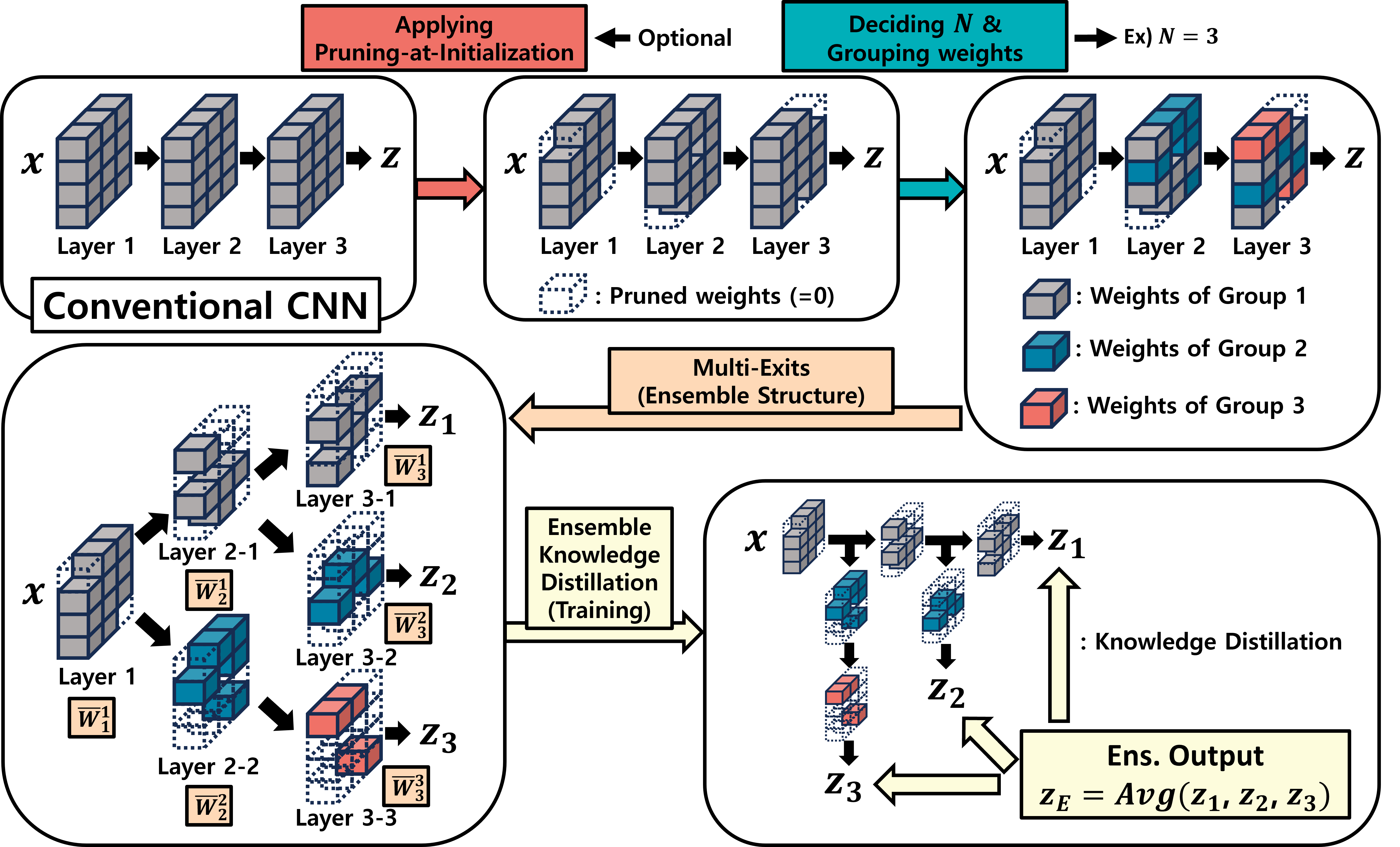}}
	\caption{Illustration of the proposed Network Fission Ensembles process. First, if the model is very large, PaI can be applied to reduce its size. Next, weight grouping is performed considering the number of ensemble members (exits) to use ($N=3$ here). Then, the network is transformed into a multi-exit structure using the grouped weights. For training, the ensemble of outputs from all exits is used as the distillation teacher to guide the learning process. Since only the computation process is changed without any explicit structural changes, no additional training and inference burden is incurred. In the illustration, we use the term `layer' in order to show an example with a simple network structure. But for popular network structures (e.g., ResNet) composed of multiple stages (each of which consists of multiple layers), weight grouping is performed in each stage rather than in each layer separately.}
	\label{Figure NFE}
\end{figure*}

\section{Related Work}
\label{sec:related}

\subsection{Ensemble Learning}
Various methods have been proposed to reduce the computational burden of ensemble learning.
There are three types of research streams for low-cost ensemble learning.

The first type uses different learning configurations to obtain differently trained models having the same structure.
An example is the method of training the same network multiple times with different initialization, which is one of the simplest forms of an ensemble \cite{ens3}.
Snapshot Ensembles (SSE) \cite{Snapshot} and Fast Geometric Ensembles (FGE) \cite{FGE} use cyclic learning rate schedules and save multiple checkpoints that correspond to ensemble members.
HyperEnsembles \cite{Hyperensemble} uses different hyperparameter configurations to obtain different models.
Since these methods need to use multiple models for ensemble inference, as the model size increases, they suffer from increased memory complexity for loading the models.

The second type applies some modification to the network itself to obtain several different outputs and construct ensembles.
TreeNet \cite{Treenet} adds additional branches in the middle of the network, but spends significant computational cost due to the increased model size.
BatchEnsemble \cite{Batchens} changes each weight matrix to the product of a shared weight and an individual rank-one matrix, introducing small additional modules. However, it has a drawback in that inference must be repeated for each ensemble member.
Multi-Input Multi-Output Ensembles (MIMO) \cite{MIMO} and its modified version, Multi-Input Massive Multi-Output (MIMMO) \cite{MIMMO}, introduce network structures that receive multiple inputs and produce multiple outputs simultaneously.
Group Ensemble \cite{GEnet} and Packed-Ensembles \cite{PACKED} use grouped convolution to obtain multiple outputs from a single model.
These methods offer the advantage of low computational overhead for the ensemble, but the diversity among ensemble members is limited because most parts of the network are shared. This leads to worse performance compared to Deep Ensembles.

The third type exploits smaller pruned sub-models for ensembles.
FreeTickets Ensembles \cite{Freeticket} obtains sub-models during the training of a main dense network by pruning. 
Prune and Tune Ensembles (PAT) \cite{PAT} obtains sub-models by pruning a trained main dense network.
However, there is a trade-off between sparsity and performance in these methods. 
As sparsity increases to improve efficiency, the number of sub-models needed to maintain performance increases, which deteriorates the low-cost advantage.

Our method overcomes the limitations of the three types of approaches. We address the limitation of the first type by using a single model, avoiding the need for multiple models. Compared to the second type, we adopt a multi-exit architecture with pruned paths, reducing the shared parts of the model and increasing the diversity of each output. To address the limitation of the third type, we train the network using ensemble knowledge distillation, which helps recover the performance loss due to the pruned structure.

\subsection{Multi-Exits}
A multi-exit structure refers to a model having one or more auxiliary classifiers (exits) in addition to the main network classifier, where parts of the network are shared among the exits.
Neural networks having multi-exits have been used for different purposes, such as anytime prediction \cite{anytime}, performance improvement \cite{BEED}, structural optimization \cite{LCT}, etc.
The multi-exit structures can be broadly categorized into two types.
The first type is a specially designed network considering the multi-output structure with a specific purpose, such as anytime prediction or low computational budget \cite{anytime}, which is less versatile.
The other type attaches auxiliary classifiers to a conventional network \cite{kd2,BEED}, which is applicable to diverse model structures.
However, in the latter approach, there is no general criterion to determine the structure of exits, and it is usually designed heuristically. 
In \cite{BEED}, it is attempted to develop a criterion for constructing the exits for performance maximization.
However, it is applicable to only certain kinds of networks.
In our method, there is no need to determine the structures of the exits a priori, which provides a more consistent and simpler solution.

\section{Proposed Method}
\label{sec:method}

\subsection{Network Fission}
Ensemble learning typically requires multiple outputs that can be combined to improve performance.
Current CNN architectures have a large number of weight parameters, some of which become redundant during training.
Considering this, we propose a method that divides the weight parameters in the network into several groups and attaches them to the network as auxiliary paths, rather than discarding them as in pruning. 
Before that, depending on the network size, we may first reduce its size using a pruning-at-initialization (PaI) method. 
Subsequently, the weight parameters are divided into multiple groups, with each group forming an auxiliary path (i.e., exit) that produces its own classification outputs. 
This results in a multi-exit structure, as illustrated in Figure \ref{Figure NFE}.
We refer to this process as Network Fission.

Let $N$ be the number of ensemble members (exits) to use. 
Define $W_i$ as the weight matrix in the $i$-th stage ($i = 1, \dots, K$) of the network. 
Before grouping the weights, the PaI mask $p_i$ can be applied to $W_i$. 
$p_i$ is a binary matrix consisting of 0s and 1s, where each value indicates whether the weight at the corresponding location is pruned (0) or unpruned (1).
The remaining weights after pruning, denoted as $\xoverline{W}_i$, are obtained by the element-wise product of $W_i$ and $p_i$:
\begin{equation} 
\label{eq1} 
\xoverline{W}_i = W_i \circ p_i. 
\end{equation}
However, this step is optional, and for simplicity, we will omit it in the remainder of this section. 
Thus, we will proceed using $W_i$ in the following explanation.

For weight grouping, the weights of the last stage $W_K$ are grouped into $N$ sets. 
To this end, we obtain a random number $j$ drawn from the categorical distribution ($1 \leq j \leq N$). 
Here, the probabilities of each outcome determine the relative sizes of the exits; we observe that it is beneficial to use the same probabilities for all outcomes for balanced weight grouping to make all exits have the same size (see Section \ref{sec:weight grouping}).
Then, the grouping mask $M^j_K$ is created, whose element is 1 if the random number is $j$ at the corresponding location and 0 otherwise.
Now, the grouped weights $W_K^j$ is obtained by the element-wise product of $W_K$ and $M^j_K$.
\begin{equation}
\label{eq2}
W_K^j = W_K \circ M^j_K.
\end{equation}
Next, the same process is performed for the weights in the penultimate stage ${W}_{K-1}$ to obtain $N-1$ weight groups; for 
${W}_{K-2}$, $N-2$ weight groups are obtained.
This process is repeated until the stage after the location at which the first exit is formed. 

The reason for reducing the number of divided groups progressively by one at each stage from the last $N$th stage is to ensure that each pathway consists of a different weight composition while retaining as many weight parameters as possible.
For example, if we divide only the last stage into $N$ groups while sharing all other stages among pathways, the outputs from the exits will exhibit large similarity, producing minimal ensemble impact.
Conversely, if we split all stages into $N$ groups, the number of weights comprising each pathway becomes too small for each exit to have enough learning capability, which results in reduced ensemble performance.
Thus, the current method is proposed to assign a sufficient number of weight parameters for each pathway and, at the same time, to maintain a certain level of diversity in outputs.

Once the grouped weights in each stage are obtained, we construct multi-exits using them.
We form the exit by combining grouped weights ${W}_i^j$ ($i=1,...,K$) having the same value of $j$.
If there is no weight corresponding to the value of $j$ in the stage, the weights with $j=1$ in that stage are used. 
For example, when $N=4$, the input-to-output paths of the four exits are as follows: 
\begin{itemize}
	\item ${W}_1^1$ $\rightarrow$ ${W}_2^1$ $\rightarrow$ ${W}_3^1$ $\rightarrow$ ${W}_4^1$
	\item ${W}_1^1$ $\rightarrow$ ${W}_2^2$ $\rightarrow$ ${W}_3^2$ $\rightarrow$ ${W}_4^2$
	\item ${W}_1^1$ $\rightarrow$ ${W}_2^1$ $\rightarrow$ ${W}_3^3$ $\rightarrow$ ${W}_4^3$
	\item ${W}_1^1$ $\rightarrow$ ${W}_2^1$ $\rightarrow$ ${W}_3^1$ $\rightarrow$ ${W}_4^4$
\end{itemize}
Using this Network Fission process, we can obtain multiple outputs for ensemble inference using a conventional single network without any additional computational cost, and ensemble inference can be performed with a much smaller amount of computation as sparsity increases.
The overall process of Network Fission is summarized in Algorithm \ref{alg_network_fission}.

Note that the network structure does not change based on the number of ways the groups are connected after weight grouping. This is because, when the weights in a stage are divided into $n$ groups, each group randomly takes $1/n$ of the weights, making the specific selection irrelevant. Therefore, in the multi-exit structure shown in Figure \ref{Figure NFE}, swapping the positions of different colored groups within a stage does not affect the performance of the network.

\begin{algorithm}[t]
	\caption{Network Fission (balanced grouping)}
	\label{alg_network_fission}
	\begin{small}
		\begin{algorithmic}
			\STATE {\bfseries Input:} $N$ ensemble members
			\STATE {\bfseries Model:} model with $K$ stages 
			\FOR{$i = K$ {\bfseries to} $2$ {\bfseries by} $-1$}
			\STATE $n \leftarrow N$  
			\FOR{$j = 1$ {\bfseries to} $n$}
			\STATE Define the total mask $\mathbf{1}$ as the union of disjoint sets $M_i^j$:
			\STATE $\mathbf{1} = \bigcup_{j=1}^{n} M_i^j, \quad M_i^s \cap M_i^t = \emptyset \quad \text{for} \quad s \neq t$
			\STATE $W_i^j \leftarrow W_i \circ M_i^j$
			\ENDFOR
			\STATE $n \leftarrow n-1$
			\IF {$n=1$}
			\STATE break
			\ENDIF
			\ENDFOR
			\STATE Form paths by combining $W_i^j$ with the same $j$
			\STATE If no weight for $j$, use $W_i^1$
		\end{algorithmic}
	\end{small}
\end{algorithm}

\subsection{Training}

One effective method for training a multi-exit network is through knowledge distillation \cite{kd2,BEED}. 
Since the main network (original path) typically has the highest performance, using it as the teacher can improve the overall network performance. 
However, in the current network obtained through network fission, no specific path has more weight parameters or a higher learning capability. 
Instead, all paths have similar learning capabilities. 
Considering this, we apply ensemble knowledge distillation, where the ensemble of all exit outputs is used as the teacher to train the network as used in \cite{BEED}.

Let $z_i$ be the logit output of the $i$-th exit in the network.
The ensemble output $z_E$ is then obtained by  
\begin{align}
\label{eqloss4}
z_E = \frac{1}{N} \sum_{i=1}^{N} z_i.
\end{align}
Using this, we compute the ensemble teacher signal $q_E$ through the softmax function with a temperature parameter $T$:
\begin{align}
\label{eqloss3}
q_E & = \text{softmax}\left(\frac{z_E}{T}\right).
\end{align}
We can also obtain the softmax output $q_i$ for each exit using same process.
Then, the training loss for our model is written as
\begin{align}
\label{eqloss2}
L & = \sum_{i=1}^{N} \bigg\{ CE(q_i, y) + \alpha KL(q_i, q_E) \bigg\},
\end{align} 
where $CE$ is the cross entropy, $KL$ is the Kullback-Leibler (KL) divergence, $y$ is the true label, and $\alpha$ is a weighting coefficient.
After training, $z_E$ is used as the ensemble output for inference.
We call our method Network Fission Ensembles (NFE).

\section{Experiments}
\label{sec:exp}
In this section, we thoroughly evaluate our proposed NFE.
First, we evalaute the accuracy and the performance of uncertainty estimation of NFE compared to state-of-the-art low-cost ensemble learning methods and Deep Ensembles.
Second, we evalaute the accuracy of NFE with respect to the sparsity of the whole network.
Third, we perform an ablation study to confirm which configuration achieves higher accuracy with respect to the PaI method and the weight grouping strategy.
Lastly, we analyze the diversity among exits and investigate how the ensemble gain is achieved in NFE.

We evaluate the methods with ResNet and Wide-ResNet \cite{wrn} for the CIFAR100 \cite{cifar} and Tiny ImageNet \cite{tinyimagenet} datasets. 
We follow the network configuration and training setting in \cite{PAT}.
Thus, for ResNet, we modify the configuration of the first convolution layer, including kernel size (7$\times$7 $\rightarrow$ 3$\times$3), stride (2 $\rightarrow$ 1), and padding (3 $\rightarrow$ 1), and remove the max-pooling operation.
For Wide-ResNet, we do not use the dropout.

Detailed training hyperparameters are as follows.
We use the stochastic gradient descent (SGD) with a momentum of 0.9 and an initial learning rate of 0.1.
The batch size is set to 128 and the maximum training epoch is set to 200 for CIFAR100 and 100 for Tiny ImageNet. 
For CIFAR100 with ResNet, the learning rate decreases by an order of magnitude after 75, 130, and 180 epochs. 
For all other cases, the learning rate is decayed by a factor of 10 at half of the total number of epochs and then linearly decreases until 90\% of the total number of epochs, so that the final learning rate is 0.01 of the initial value.
The L2 regularization is used with a fixed constant of $5 \times 10^{-4}$. 
$\alpha$ in Eq. (\ref{eqloss2}) is set to 1.
The temperature ($T$) in Eq. (\ref{eqloss3}) is set to 3, a commonly used value in image classification studies.
For Tiny ImageNet, we initialize the network using the model pre-trained for CIFAR100.
All experiments are performed using Pytorch with NVIDIA RTX 8000 graphics processing units (GPUs).
We conduct all experiments three times with different random seeds and report the average results.

\subsection{Performance Comparison}  

We compare the performance of NFE with Deep Ensembles \cite{ens3} and the other state-of-the-art low-cost ensemble learning methods, including SSE \cite{Snapshot}, FGE \cite{FGE}, PAT \cite{PAT}, GENet \cite{GEnet}, TreeNet \cite{Treenet}, BatchEnsemble \cite{Batchens}, MIMO \cite{MIMO}, FreeTickets (Dynamic Sparse Training (DST), Efficient Dynamic Sparse Training (EDST)) \cite{Freeticket}, and Packed-Ensembles \cite{PACKED}.
Following \cite{MIMO,Freeticket,PAT}, we compare the accuracy, negative log-likelihood (NLL), and expected calibration error (ECE) for CIFAR10 and CIFAR100 when the initial network structure is Wide-ResNet28-10.
We also report the relative ratio of the total number of floating point operations (FLOPs) for inference compared to the single model.

The results for CIFAR-100 are shown in Table \ref{table5}, while the results for CIFAR-10 are provided in the supplementary material. 
Figure \ref{Figure accflops} visually compares different methods in terms of accuracy vs. computational cost. 
Since Packed-Ensembles uses CutMix \cite{cutmix} data augmentation, it is distinguished separately in Table \ref{table5}. 
To ensure a fair comparison, we also evaluate NFE with CutMix applied, but the results with CutMix are excluded from being the best results.
To demonstrate the performance of NFE across various cases, we show the results when the number of ensemble members is two and three, both without PaI (\emph{S}=0) and with PaI (\emph{S}=0.5), where SNIP \cite{SNIP} is used for PaI.

Our NFE achieves significantly enhanced accuracy compared to Deep Ensembles and the other state-of-the-art low-cost ensemble methods for both datasets.
Moreover, despite drastically reduced FLOPs by pruning, our method outperforms the best-performing existing methods. 
Even without pruning, our method shows almost no increase in FLOPs and does not require additional training epochs beyond those needed for the single model.
In terms of NLL, our method achieves a satisfactory result with only a slight difference of 0.015 (corresponding to a relative difference of 2.37\%) from the best.
Although the ECE performance of our method is slightly lower compared to the best performance, our method is still comparable to most of the compared methods.
Even when compared with Packed-Ensembles with CutMix, NFE with CutMix significantly outperforms in terms of accuracy, NLL, and ECE.

\begin{table}[t] 
	\footnotesize
	\begin{center}     
		{\caption{Performance comparison for CIFAR100 with Wide-ResNet28-10. We mark the best ensemble results in bold and the second-best results with underlines. The results of the methods marked with * and ** are from \cite{PAT} and \cite{PACKED}, respectively. \emph{S} means sparsity. Packed-Ensembles is evaluated separately due to the use of data augmentation.}
			\label{table5}}
		\begin{tabular}{lcccccccc}
			\hline
			\rule{0pt}{12pt}
			Method &Acc (\%) $\uparrow$&NLL $\downarrow$&ECE $\downarrow$& FLOPs $\downarrow$
			\\
			\hline
			\\[-6pt]
			Single Model& 81.5 & 0.741 & 0.047 &  3.6$\times10^{17}$ \\
			SSE ($N$=5)*& 82.1 & 0.661 & 0.040 &  1.00x \\
			FGE ($N$=12)*& 82.3 & 0.653 & 0.038 & 1.00x \\
			PAT ($N$=6)*& 82.7 & \underline{0.634} & \textbf{0.013} & 0.85x \\
			TreeNet ($N$=3) & 82.5 & 0.681 & 0.043 & 2.53x \\
			GENet ($N$=3) & 82.7 & 0.707 & 0.051 & 0.94x \\
			MIMO ($N$=3)*& 82.0 & 0.690 & \underline{0.022} & 1.00x \\
			EDST ($N$=7)*& 82.6 & 0.653 & 0.036 &  0.57x \\
			DST ($N$=3)* & 82.8 & \textbf{0.633} & 0.026 & 1.01x \\
			BatchEnsemble ($N$=4)**& 82.3 & 0.835 & 0.130 & 3.99x\\
			Deep Ensembles ($N$=3) & 83.0 & 0.676 & 0.037 & 3.00x \\
			NFE ($N$=2, \emph{S}=0) & \textbf{83.5} & 0.656 & 0.044 & 1.01x \\
			NFE ($N$=2, \emph{S}=0.5) & \underline{83.4} & 0.649 & 0.039 & 0.52x \\
			NFE ($N$=3, \emph{S}=0) & \textbf{83.5} & 0.658 & 0.061 &  1.02x \\
			NFE ($N$=3, \emph{S}=0.5) & 83.1 & 0.669 & 0.045 & 0.53x \\
			\hline
			\makecell[l]{Packed-Ensembles \\ with CutMix ($N$=4)** }  & 83.9 & 0.678 & 0.089 & 1.00x &  \\	
			\makecell[l]{NFE ($N$=3, \emph{S}=0) \\ with CutMix} & 85.1 & 0.577 & 0.020 & 1.02x \\
			\hline
		\end{tabular}
	\end{center}
\end{table}

\begin{figure}[t] 
	\centerline{\includegraphics[width=0.65\textwidth]{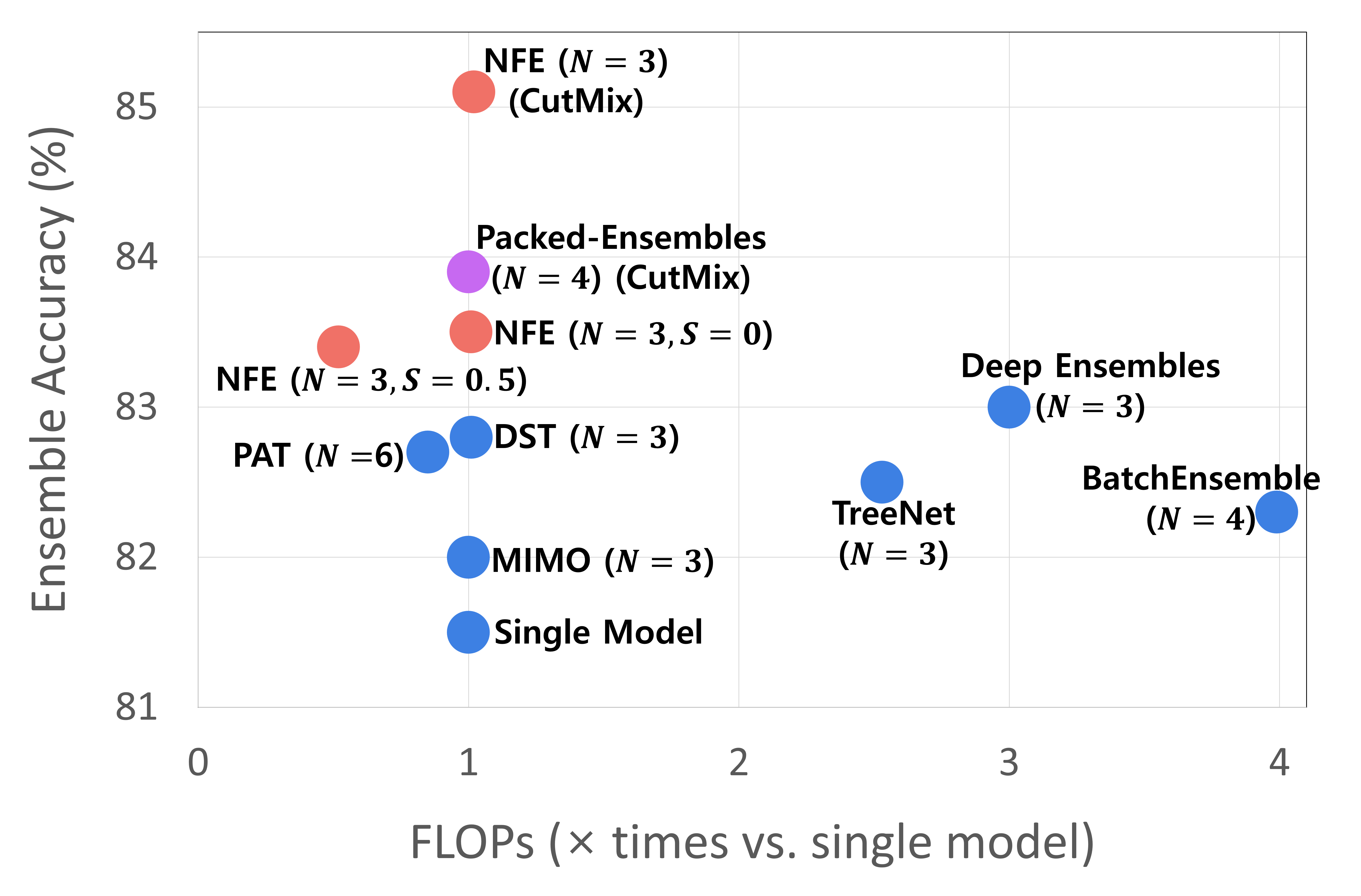}}
	\caption{Ensemble accuracy (\%) vs. FLOPs for inference for CIFAR100 with Wide-ResNet28-10.}
	\label{Figure accflops}
\end{figure}

\subsection{Sparsity vs. Performance}

We evaluate the performance of NFE with respect to the sparsity of the network.
Although ensemble inference is an effective approach to improve performance, it is desirable to reduce the computational burden further to use it over a wide range of computational environments.
Even though NFE does not introduce additional computation compared to the single network, we show that NFE can achieve high performance even when the sparsity increases by SNIP.

Tables \ref{table0} and \ref{table2} show the test accuracy for CIFAR100, when ResNet18 and WRN28-10 are used as the initial network, respectively.
Table \ref{table3} shows the test accuracy for Tiny ImageNet, when WRN28-10 is used as the initial network.
We consider four sparsity levels (0, 0.25, 0.5, 0.75).
In the tables, ``Single'' denotes the case when the given network structure is used without Network Fission.
TreeNet \cite{Treenet} is included in our comparison since it is similar to our NFE in that it also performs ensemble learning by constructing multi-exits.
In order to examine the effectiveness of NFE depending on the number of ensemble members (i.e., $N$), the tables present the results of NFE with the minimum and maximum numbers of members for each initial network structure.
When the number of ensemble members is minimum ($N$=2), we use \textit{Res*2*4} and \textit{WRN1*3} for each network, respectively, because this setting achieves better performance.

In the tables, our NFE achieves significantly enhanced accuracy compared to the single model and TreeNet for both datasets.
When the sparsity increases, the improvement by ensembles is maintained.
Even when a half of the weights in the network are pruned, we can still obtain performance improvement by about 2\% for CIFAR100 and 3\% for Tiny ImageNet compared to the single model.
This improvement can be attributed largely to the training loss.
Although weight grouping causes partial loss of the weights in each path, distillation using the ensemble teacher can compensate for this deficiency.
When the maximum possible number of ensemble members is used in NFE ($N$=3), the accuracy is slightly lowered. 
This is because the Network Fission process is repeated more for a larger $N$ and the number of weights in each path decreases. 
The performance decline is larger for a higher sparsity. 
Nevertheless, NFE outperforms both the single model and TreeNet in most cases.

\subsection{Ablation Study}
We investigate the difference in performance with respect to different types of PaI methods and weight grouping strategies.
Through this, we examine how different pruning methods and weight grouping strategies affect performance by evaluating the performance of NFE using ResNet18 trained with only the cross-entropy loss for CIFAR100.

\subsubsection{Pruning-at-Initialization}

To evaluate the performance with respect to the PaI method, we compare four PaI methods, SNIP \cite{SNIP}, GraSP \cite{Grasp}, Erd\H{o}s-R\'enyi-Kernel (ERK) \cite{Unreason}, and SynFlow \cite{Synflow}.
As a criterion for weight pruning, SNIP and GraSP use the gradients of the training loss with respect to the weights, ERK uses a random topology with higher sparsity in larger layers, and SynFlow uses the magnitudes of the weights considering the adjacent layers' weights.
The pruning mask is generated by each method, and the performance of NFE having four exits is compared. 
When applying the pruning mask, we exclude the last fully-connected layer as done in \cite{Unreason}, and also the convolution layers used for down-sampling in the residual path.
These layers are highly small in size compared to the whole network, but greatly significant for performance.
Thus, we do not perform pruning for these layers in all experiments.

\begin{table}
	\footnotesize
	\begin{center}
		{\caption{Comparison of the test accuracy for CIFAR100 with respect to the sparsity when ResNet18 is used as the initial network structure. The best performance at each sparsity level is marked in bold.}
			\label{table0}}
		\begin{tabular}{ccccccccc}
			\hline
			\rule{0pt}{12pt}
			&\multicolumn{4}{c}{Accuracy (\%)}\\ 
			\cline{2-5}
			\rule{0pt}{12pt}
			Sparsity & Single & TreeNet ($N$=4) & NFE ($N$=2) & NFE ($N$=4)\\
			\hline
			\\[-6pt]
			0& 77.64 & 79.41 & \textbf{80.68} & 79.73 \\
			0.25& 77.44 & 79.16 & \textbf{80.54} & 79.59 \\
			0.50& 77.36 & 78.59 & \textbf{80.01} & 79.14 \\
			0.75& 76.41 & 77.64 & \textbf{79.13} & 77.62 \\
			\hline
		\end{tabular}
	\end{center}
\end{table}

\begin{table}
	\footnotesize
	\begin{center}
		{\caption{Comparison of the test accuracy for CIFAR100 with respect to the sparsity when Wide-ResNet28-10 is used as the initial network structure. The best performance at each sparsity level is marked in bold.}
			\label{table2}}
		\begin{tabular}{ccccccccc}
			\hline
			\rule{0pt}{12pt}
			&\multicolumn{4}{c}{Accuracy (\%)}\\ 
			\cline{2-5}
			\rule{0pt}{12pt}
			Sparsity & Single & TreeNet ($N$=3) & NFE ($N$=2) & NFE ($N$=3) \\
			\hline
			\\[-6pt]
			0& 81.5 & 82.5 & \textbf{83.5} & \textbf{83.5} \\
			0.25& 81.4 & 81.9 & \textbf{83.4} & 83.2 \\
			0.50& 81.3 & 81.1 & \textbf{83.4} & 83.1 \\
			0.75& 80.9 & 80.5 & \textbf{83.0} & 80.7 \\
			\hline
		\end{tabular}
	\end{center}
\end{table}

\begin{table}
	\footnotesize
	\begin{center}
		{\caption{Comparison of the test accuracy for Tiny ImageNet with respect to the sparsity when Wide-ResNet28-10 is used as the initial network structure. The best performance at each sparsity level is marked in bold.}
			\label{table3}}
		\begin{tabular}{ccccccccc}
			\hline
			\rule{0pt}{12pt}
			&\multicolumn{4}{c}{Accuracy (\%)}\\ 
			\cline{2-5}
			\rule{0pt}{12pt}
			Sparsity & Single & TreeNet ($N$=3) & NFE ($N$=2) & NFE ($N$=3) \\
			\hline
			\\[-6pt]
			0& 68.2 & 69.0 & \textbf{71.0} & 70.6 \\
			0.25& 67.9 & 68.5 & \textbf{71.0} & 70.2 \\
			0.50& 67.4 & 66.9 & \textbf{70.9} & 67.7 \\
			0.75& 65.0 & 65.3 & \textbf{68.4} & 66.6 \\
			\hline
		\end{tabular}
	\end{center}
\end{table}

\begin{figure}[t] 
	\centerline{\includegraphics[width=0.6\textwidth]{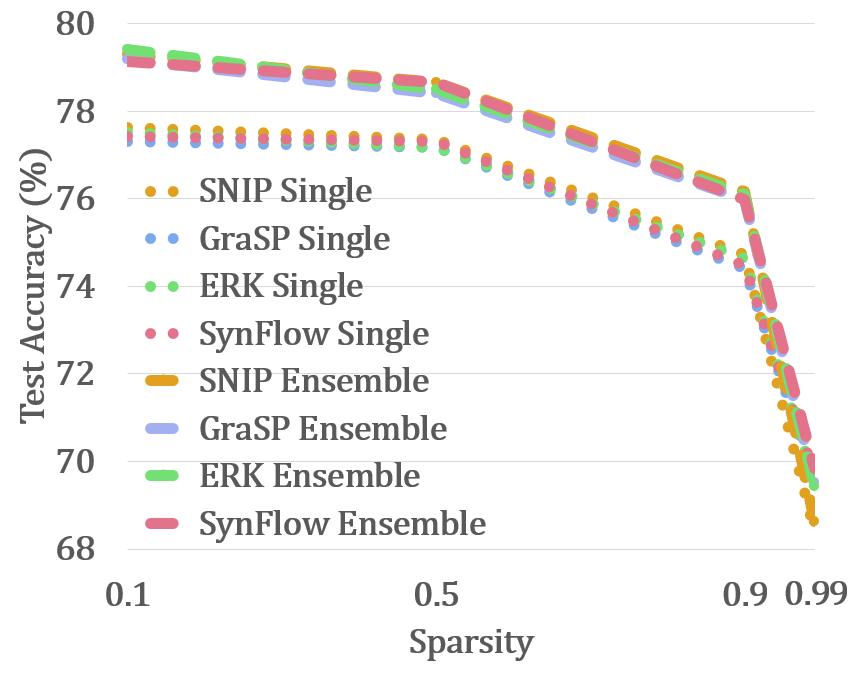}}
	\caption{Test accuracy (\%) of different PaI methods with ResNet18 for CIFAR100 with respect to the sparsity.}
	\label{Figure PaI comparison}
\end{figure}

The results are shown in Figure \ref{Figure PaI comparison}. 
We compare the accuracy achieved from the single main network path and the ensemble of all exits.
We evaluate the four cases of sparsity: 0.1, 0.5, 0.9, and 0.99.
The results show that there is no significant difference in the accuracy of both the main network and the ensemble among the PaI methods.
When we extremely increase the sparsity (i.e., 0.99), differences among the methods become noticeable.
However, the performance degradation is severe in all PaI methods, so this case does not have practical usage for ensemble learning.
When the sparsity is below 0.99, we can get the benefit of a multi-exit ensemble.
Since all the methods perform similarly in the range of sparsity (from 0.0 to 0.9) we are interested in, we can choose the computationally simplest one, i.e., SNIP, as the pruning method if we opt to apply the PaI method.

\subsubsection{Weight Grouping}
\label{sec:weight grouping}

For the Network Fission process, it is necessary to assign the proportion of weights to each exit. 
We compare the accuracy by varying the weight grouping ratio for each exit.
To simplify the procedure, we use the multi-exits having only two exits without PaI.
In this case, three kinds of network are possible, where one exit corresponds to the blue path (original main network) and the other (auxiliary exit) is one among the green, orange, and purple paths in Figure \ref{Figure NFE}.
We denote each network as \textit{Res1**4}, \textit{Res*2*4}, and \textit{Res**34}, respectively.

\begin{figure}[t] 
	\centerline{\includegraphics[width=0.6\textwidth]{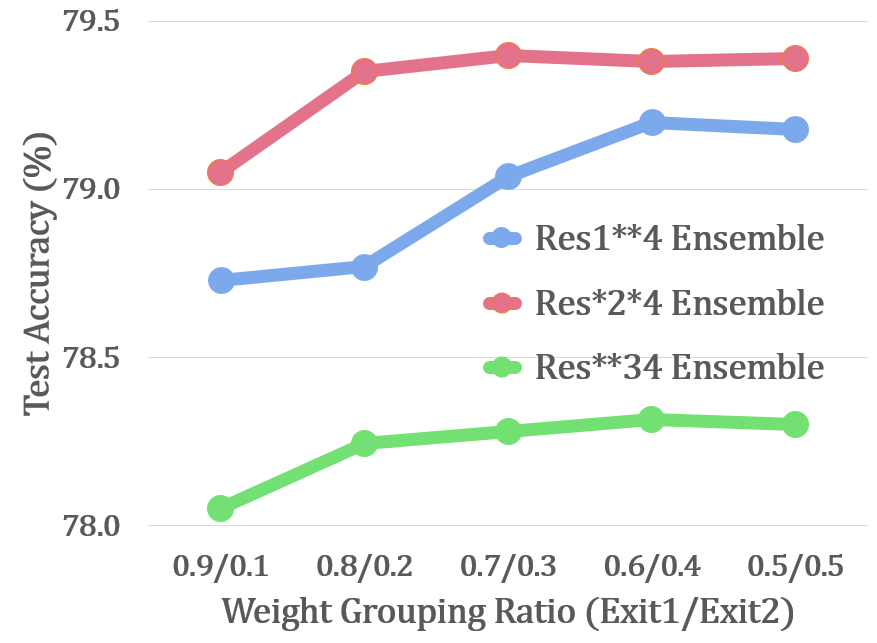}}
	\caption{Test accuracy (\%) for different weight grouping ratios with ResNet18 for CIFAR100.}
	\label{Figure weight grouping}
\end{figure}

Figure \ref{Figure weight grouping} shows that the accuracy increases as the grouping ratios get more similar. 
When the proportions of weights in the two exits are largely unbalanced and many weights are assigned to one exit, the exit with a smaller number of weights performs poor, which contributes to the ensemble inference negligibly or even adversely.
Thus, the balanced weight grouping strategy is advantageous by distributing the weights equally to all exits to prevent performance degradation of any exit and improve ensemble performance.
Therefore, the balanced weight grouping strategy is an effective choice for NFE.

\begin{table}
	\footnotesize
	\begin{center}
		{\caption{Prediction disagreement (PD) and cosine similarity (CS) between two ensemble members ($N$=2), and accuracy (\%) of each ensemble member and the ensemble for CIFAR100 with Wide-ResNet28-10. \emph{S} means sparsity.}
			\label{table7}}
		\begin{tabular}{lcccccccc}
			\hline
			\rule{0pt}{12pt}
			Method & PD & CS & Acc 1 & Acc 2 & Acc Ens \\
			\hline
			\\[-6pt]
			TreeNet & 0.102 & 0.947 & 81.3 & 81.2 & 82.4 \\
			Deep Ensembles & 0.153 & 0.904 & 81.6 & 81.5 & 82.9 \\
			NFE (CE) (\emph{S}=0) & 0.154 & 0.906 & 80.9 & 80.8 & 82.6 \\
			NFE (CE) (\emph{S}=0.5) & 0.157 & 0.898 & 80.8 & 80.5 & 82.5 \\
			NFE (CE+KL) (\emph{S}=0) & 0.104 & 0.941 & 82.8 & 82.7 & 83.5 \\
			NFE (CE+KL) (\emph{S}=0.5) & 0.097 & 0.949 & 82.7 & 82.6 & 83.4 \\
			\hline
		\end{tabular}
	\end{center}
\end{table}

\subsection{Diversity Analysis}

To analyze the performance improvement of NFE, we evaluate the diversity of the exits in the NFE structure.
As in \cite{Freeticket,BEED}, we measure the pairwise diversity among ensemble members using two metrics: prediction disagreement and cosine similarity.
The prediction disagreement is defined as the ratio of the number of test samples that two ensemble members classify differently \cite{divdis}.
The cosine similarity is measured between the softmax outputs of two ensemble members \cite{divcos}. 
The results for CIFAR-100 with $N$=2 are shown in Table \ref{table7}, while the results for CIFAR-10 are provided in the supplementary materials.

We note that the training loss directly affects the diversity in NFE. 
Our training loss in Eq. (\ref{eqloss2}) includes the KL divergence (in addition to CE), which performs knowledge distillation for all exits using the same ensemble teacher. 
This would reduce the diversity among the exits compared to the case where only CE is used, which can be observed in the tables. 
Nevertheless, the case using both CE and KL shows higher ensemble accuracy than the case using only CE. 
This is because the performance of each ensemble member is greatly improved in the former case, which can be also observed in the tables. 
This brings an additional advantage of NFE that, even when the ensemble inference is not feasible due to certain computational constraints, it is still possible to obtain high classification accuracy using only one exit.

\section{Conclusion}
\label{sec:conclusion}
We proposed a new approach to ensemble learning with no additional cost, which converts the network structure into a multi-exit configuration.
By grouping weights and creating auxiliary paths, our NFE forms a multi-exit structure that leverages ensemble knowledge distillation, enabling effective training of the sparsified network.
It was demonstrated that compared to existing low-cost ensemble learning methods, our method achieves the highest accuracy.

We believe that our NFE method can be effectively adopted in real-world applications where high classification performance is desired but computational resources are limited. Furthermore, NFE is not only applicable to image classification but can also be extended to other computer vision tasks such as object detection and semantic segmentation. However, due to its nature of splitting the network, increasing the number of exits ($N$) may be somewhat limited, which can hinder scalability. Further research on this issue would be desirable. Additionally, as part of our future work, we are exploring efficient implementation of the grouped weights, which, once achieved, will allow researchers to apply NFE to CNNs more easily.

\section*{Acknowledgement}
This work was supported by the National Research Foundation of Korea (NRF) grant funded by the Korea government (MSIT) (No. RS-2024-00453301) and by the Yonsei Signature Research Cluster Program of 2024 (2024-22-0161).

\bibliographystyle{plain}
\bibliography{egbib}

\end{document}